# ENTROPY, PERCEPTION, AND RELATIVITY

Stefan Jaeger

Language and Media Processing Laboratory
Institute for Advanced Computer Studies
University of Maryland
College Park, MD 20742-3275
*jaeger@umiacs.umd.edu*

## Abstract

In this paper, I expand Shannon's definition of entropy into a new form of entropy that allows integration of information from different random events. Shannon's notion of entropy is a special case of my more general definition of entropy. I define probability using a so-called performance function, which is de facto an exponential distribution. Assuming that my general notion of entropy reflects the true uncertainty about a probabilistic event, I understand that our perceived uncertainty differs. I claim that our perception is the result of two opposing forces similar to the two famous antagonists in Chinese philosophy: Yin and Yang. Based on this idea, I show that our perceived uncertainty matches the true uncertainty in points determined by the golden ratio. I demonstrate that the well-known sigmoid function, which we typically employ in artificial neural networks as a non-linear threshold function, describes the actual performance. Furthermore, I provide a motivation for the time dilation in Einstein's Special Relativity, basically claiming that although time dilation conforms with our perception, it does not correspond to reality. At the end of the paper, I show how to apply this theoretical framework to practical applications. I present recognition rates for a pattern recognition problem, and also propose a network architecture that can take advantage of general entropy to solve complex decision problems.

The support of this research by the Department of Defense under contract MDA-9040-2C-0406 is gratefully acknowledged.

# 1 Introduction

Uncertainty is our constant companion in everyday's decision making. Being able to deal with uncertainty is thus an essential requirement for intelligent behavior in real-world environments. Naturally, knowing the exact amount of uncertainty involved in a particular decision is a very useful information to have. Mathematically, the classic way of measuring the uncertainty for a random event is to compute its information based on the definition of entropy introduced by Shannon [15]. In this paper, however, I introduce a new, general form of entropy that is motivated by my earlier work on classifier combination [3, 4, 6]. The idea of classifier combination, or sensor fusion in general, is to combine the outcomes of different sub-optimal processes into one integrated result. Ideally, the integrated process performs better in the given application domain than each individual process alone. In order to integrate different processes into a single process, computers need to deal with the uncertainties involved in the outcomes of each individual process. For classifier combination, several combination schemes have already been suggested. The current state-of-the-art, however, has not given its final verdict on this issue yet. In my earlier work, I proposed an informational-theoretical approach to this problem. The main idea of this approach is to normalize confidence values in such a way that their nominal values match their conveyed information, which I measure on a training set in the application domain. The overall combined confidence for each class is then simply the sum of the normalized confidence values of each individual classifier. In this paper, I am going to elaborate on my earlier ideas by looking at them from the general entropy's point of view.

I structured the paper as follows: Following this introduction, Section 2 repeats the definition of entropy as introduced by Shannon, and compares it to my new and more general definition. Section 3 provides a short introduction into my earlier work on informational confidence and repeats the main postulates and their immediate consequences. Section 4 describes how I understand confidence as the result of an interplay of two opposing forces. In Section 5, this insight will show the sigmoid function of classic back-propagation networks in a different light, namely as a kind of mediator between these two forces. A closer inspection in Section 6 reveals that the net effect of both opposing forces equals one single force in points defined by the golden ratio. In Section 7, I relate the introduced forces to the well-known forces of Yin and Yang in Chinese philosophy. In particular, I show how we can derive the typical Yin-Yang symbol from the assumptions made. In Section 8, I explore common grounds of the general framework presented here and Einstein's Special Relativity. I provide an interesting motivation for the time dilation in Einstein's Special Relativity. Section 9 is then going to show how we can learn informational confidence values, illustrating the learning process with a practical example of handwritten Japanese character recognition. This section also proposes a network architecture for learning based on the ideas introduced in the previous sections. Finally, a summary with the main results concludes the paper.

## 2  Entropy

Entropy is a measure for the uncertainty in a random event or signal. Alternatively, we can understand entropy as the amount of information conveyed by the random event or carried by the signal. Entropy is a general concept that has applications in statistical mechanics, thermodynamics, and of course information theory in computer science. The latter will be the focus of my attention in the following. At the end of my paper, I will present an interesting connection with Einstein's Special Relativity and physics, though.

Claude E. Shannon introduced entropy as a measure for randomness in his 1948 seminal paper "A Mathematical Theory of Communication." For a discrete random event with $n$ possible outcomes, Shannon defines the entropy $H$ as the sum of expected information $K_i$ for each outcome $i$:

$$H = \sum_{i=1}^{n} K_i \qquad (1)$$

Shannon uses the negative logarithm to compute information itself. In this way, he can simply add the information of two independent outcomes to get the combined information of both. Accordingly, each $K_i$ in (1) reads as follows:

$$K_i = -p(i) \ln(p(i)), \qquad (2)$$

with $p(i)$ denoting the probability of the i-th outcome. The entropy reaches a maximum when all $p(i)$ are equal, which indicates maximum uncertainty. On the other hand, the entropy is minimal; i.e. zero, if exactly one $p(i)$ is 1 and all other outcomes have a probability of zero.

I am now introducing the following more general variant that I will be using instead of (2) to compute the entropy $H$:

$$K_i = -p(K_i) \ln(p(K_i)) \qquad (3)$$

In this new form, the expected information for each outcome appears on both sides of the equation, effectively making (3) a fixed point equation. Instead of using the probability $p(i)$ of an outcome, I am now using the probability of the outcome's specific information. I also do not require the sum of all probabilities $p(K_i)$ to be one.

A straightforward comparison of (2) and (3) shows that Shannon's definition of entropy and its more general variant are the same if each outcome satisfies the following equation:

$$p(i) = p(K_i) \qquad (4)$$

In other words, both definitions of entropy are the same when the probability of each outcome matches the probability of its information, which we can consider to be a fixed point.

The next section gives a motivation for the general entropy formula using pattern recognition, and in particular classifier combination, as a practical example.

## 3  Informational Confidence

Pattern recognition is a research field in computer science dealing with the automatic classification of pattern samples into different classes. Depending on the application domain, typical classes are; e.g., characters, gestures, traffic signs, faces, etc. For a given unknown test pattern, most classifiers return both the actual classification result in form of a ranked list of class labels, and corresponding values indicating the confidence of the classifier in each class label. I will be using the term "confidence value" for these values throughout the paper, but I should mention that other researchers may prefer different terms, such as "score" or "likelihood." In practical classifier systems, confidence values are usually only rough approximations of their mathematically correct values. In particular, they very often do not meet the requirements of probabilities. While this usually does not hamper the operation of a single classifier, which only depends on the relative proportion of confidence values, it causes problems in multiple classifier systems, which need the proper values for combination purposes. Post-processing steps, such as linguistic context analysis for character recognition, can also benefit from more accurate confidence values.

Combination of different classifiers in a multiple classifier systems has turned out to be a powerful tool for reducing the uncertainty involved in a classification problem [8]. Researchers have shown in numerous experiments that the performance of the combined classifiers can outperform the performance of each single classifier. Nevertheless, researchers are still undecided about how to best integrate the confidence values of each individual classifier into one single confidence. In earlier work, I proposed so-called informational confidence values as as a way to combine multiple confidences values [3, 4, 6]. The idea of informational confidence values is to introduce a standard of comparison allowing fair comparison and easy integration of confidence values generated by different classifiers. The definition of informational confidence values relies on two central postulates:

1  Confidence is information

2  Information depends on performance

The first postulate states that each confidence value conveys information, and it consequently requires that the nominal value of each confidence value should equal the information conveyed. The second postulate then logically continues by requiring that the amount of information conveyed should depend on the performance of the confidence value in the application domain. From both postulates taken together, I can follow that confidence depends on performance via information. To formalize these requirements, let me assume that each classifier $C$ can output confidence values from a set of confidence values $K_C$, with

$$K_C = \{K_0, K_1, \ldots, K_i, \ldots, K_N\} \quad (5)$$

Let me further assume that $K_N$ indicates the highest confidence classifier $C$ can output. The following fixed point equation then defines a linear relationship between confidence and information, with the latter depending on the performance complement of each confidence value.

$$K_i = E * I\left(\overline{p}(K_i)\right) + C \tag{6}$$

We see that the confidence values $K_i$ appear on both sides of Equation (6), essentially making it a fixed point equation with the so-called informational confidence values as fixed points. Using the performance complement ensures that higher confidence values with better performance convey more information than lower confidence values when we apply Claude Shannon's logarithmic notion of information [15]. According to Shannon, information of a probabilistic event is the negative logarithm of its probability. More information on Shannon's work and the implications of his strikingly simple definition of information can be found in [13, 14, 16].

By setting constant $C$ to zero, inserting the negative logarithm as information function $I$, and using $1 - p(K_i)$ as performance complement, I simplify Equation (6) to the following definition of informational confidence:

$$K_i = -E * \ln\left(1 - p(K_i)\right) \tag{7}$$

The still unknown parameters necessary to compute informational confidence values according to (7) are $E$ and $p(K_i)$. A straightforward transformation of (7) sheds more light on these two parameters:

$$K_i = -E * \ln\left(1 - p(K_i)\right)$$
$$\iff e^{-\frac{K_i}{E}} = 1 - p(K_i)$$
$$\iff p(K_i) = 1 - e^{-\frac{K_i}{E}} \tag{8}$$

The result shows that the performance function $p(K_i)$ describes an exponential distribution with expectation value $E$. This follows from the general definition of an exponential density function $e_\lambda(x)$ with parameter $\lambda$:

$$e_\lambda(x) = \begin{cases} \lambda * e^{-\lambda x} & : \ x \geq 0 \\ 0 & : \ x < 0 \end{cases} \quad \lambda > 0 \tag{9}$$

For each $\lambda$, the enclosed area of the density function equals 1:

$$\int_{-\infty}^{\infty} e_\lambda(x)\, dx = \int_0^{\infty} \lambda * e^{-\lambda x}\, dx = 1 \quad \forall \lambda > 0 \tag{10}$$

Figure 1 shows three different exponential densities differing in their parameter $\lambda$, with $\lambda = 100$, $\lambda = 20$, and $\lambda = 10$ respectively. The parameter $\lambda$ has a direct influence on the steepness of the exponential density function. The higher $\lambda$, the steeper the density function.

The corresponding distribution $E_\lambda(k)$, which describes the probability that the random variable assumes values lower than or equal to a given value $k$, computes as follows:

$$E_\lambda(k) = \int_{-\infty}^{k} e_\lambda(x)\, dx$$

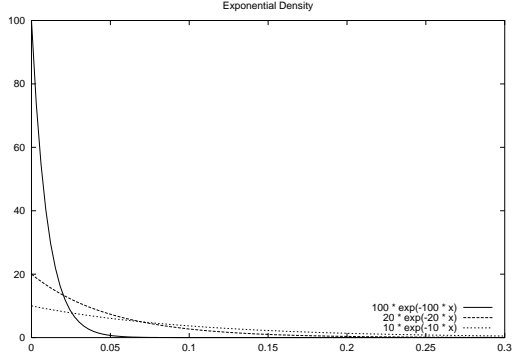

Figure 1: Exponential density for $\lambda = 100$, $\lambda = 20$, and $\lambda = 10$.

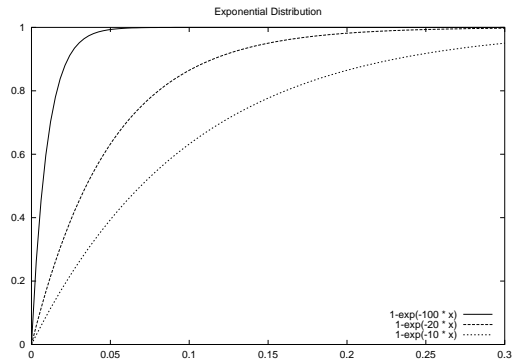

Figure 2: Exponential distribution for $\lambda = 100$, $\lambda = 20$, and $\lambda = 10$.

$$
\begin{aligned}
&= \int_0^k \lambda * e^{-\lambda x}\, dx \\
&= \left[-e^{-\lambda x}\right]_0^k \\
&= 1 - e^{-\lambda k}
\end{aligned}
\qquad (11)
$$

Figure 2 shows the distributions for the three different densities depicted in Figure 1, with $\lambda = 100$, $\lambda = 20$, and $\lambda = 10$. The parameter $\lambda$ influences again the steepness: A larger $\lambda$ entails a steeper distribution. For each parameter $\lambda$, the distribution function converges on 1 with increasing confidence. Another important feature is the relationship between parameter $\lambda$ and the expectation value $E$ of the exponentially distributed random variable. Both are in inverse proportion to each other, with $E = \frac{1}{\lambda}$. Accordingly, the expectation values corresponding to the exponential densities in Figure 1, and distributions in Figure 2, are $E = \frac{1}{100}$, $E = \frac{1}{20}$, and $E = \frac{1}{10}$, respectively.

When we compare the performance specification in (8) with the exponential distribution in (11), we see that the only difference lies in the exponent of the exponential function. In fact, performance function and exponential distribution become identical for $\lambda = \frac{1}{E}$. This result shows that the performance function $p(K_i)$ describes the distribution of exponentially distributed confidence values with expectation $E$. We can therefore consider confidence as an exponentially distributed random variable with parameter $\lambda = \frac{1}{E}$.

The performance theorem summarizes this important result:

**Performance Theorem:**
A classifier $C$ with performance $p(K)$ provides informational confidence $K = -E * \ln(1 - p(K))$ if, and only if, $p(K)$ is an exponential distribution with expectation $E$.

The performance theorem explains the meaning and implications of the parameters $E$ and $p(K)$. For classifiers violating the performance theorem, the equation stated in the performance theorem allows to compute the proper informational confidence values as long as we know the specific values of $E$ and $p(K)$. Section 9 will later show how we can estimate these parameters on a given evaluation set.

In the next section, I take the idea of informational confidence a step further and introduce a second type of confidence called counter-confidence, which describes the confidence of the classifier in the falseness of its output. The subsequent sections then elaborate on this concept and present new theoretical results and discuss their implications.

## 4 Opposing Forces

I am assuming that decision making is based on two opposing forces, one supporting a certain outcome and one arguing against it. In particular, I am going to propose a formalization of both forces, which I name Force A and Force B for the time being, based on the fixed point equation of the performance theorem. In fact, I postulate that Force A is already defined by this equation. Force B only differs in its interpretation of performance.

### 4.1 Force A

The first force, Force A, describes the confidence in a particular decision. Accordingly, I use the fixed point equation of informational confidence values as the definition of Force A:

$$K = -E * \ln(1 - p(K)) \quad (12)$$

The left-hand side of this equation denotes the magnitude of Force A. It is the product of information in the Shannon sense and an expectation value in the statistical sense. As shown above, the performance function $p(K)$ follows immediately as $p(K) = 1 - e^{-\frac{K}{E}}$. If the performance in the logarithmic expression on the right-hand side of (12) is 1, and the expectation $E$ is positive, then A-Force becomes infinity. On the other hand, if the performance is zero, then the logarithm becomes zero and there is no A-force at all.

### 4.2 Force B

The second force, Force B, is defined similarly but performs complementary to Force A. Force B describes information that depends directly on the performance and not on the performance complement. Accordingly, the following modified fixed point equation describes Force B:

$$K = -E * \ln(p(K)) \quad (13)$$

The difference to Force A lies in the interpretation of the performance function $p(K)$, which follows again from a straightforward transformation:

$$K = -E * \ln(p(K))$$
$$\iff p(K) = e^{-\frac{K}{E}} \tag{14}$$

We see that the performance function of Force B is similar to the performance of Force A. However, it looks at the problem from a different side. Instead of describing the area delimited by $K$ under the exponential density curve, it describes the remaining area that is not delimited. Parameter E is again a statistical expectation value. Unlike Force A, Force B becomes infinity for a performance equal to zero and positive expectation. It becomes zero whenever the performance is perfect, i.e. $p(K) = 1$. While Force A defines informational confidence values, Force B can be considered as defining informational counter-confidence values.

### 4.3 Interplay of Forces

Having defined both Force A and Force B, I postulate that all decision processes are the result of the interplay between these two forces. What we can actually experience when making decisions is the dominance one of these forces has achieved over its counterpart. Mathematically, I understand that this dominance is the net effect of both forces and thus use the difference between the defining equations in (12) and (13) to describe it:

$$K = -E * \ln\left(\frac{1 - p(K)}{p(K)}\right) \tag{15}$$

This equation is a fixed point equation itself. It describes the net force, which is the result of both forces acting simultaneously. Naturally, the net force becomes zero when Force A equals Force B. This is the case when either the expectation value is zero or the performance $p(K)$ is 0.5. The net force becomes either infinity or minus infinity when one force dominates completely over its counterpart. In particular, the net force becomes infinity when Force A dominates with $p(K) = 1$ and minus infinity when Force B dominates with $p(K) = 0$.

The following two sections are going to present two more interesting theoretical results, which are a direct consequence of the net force defined by (15), namely the sigmoid function and the golden ratio. Section 7 will later relate Force A and Force B to the well-known antagonistic forces in Chinese philosophy: Yin and Yang.

### 5 Sigmoid Function

A closer look at the net force defined in (15) reveals that the performance function is indeed a well-known function. A straightforward derivation leads to the following result:

$$K = -E * \ln\left(\frac{1 - p(K)}{p(K)}\right)$$

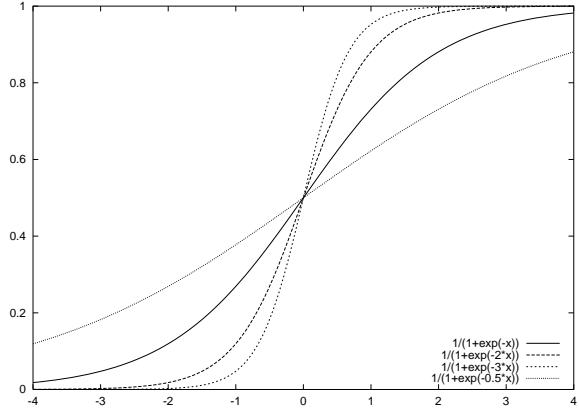

Figure 3: Sigmoid function.

$$\iff e^{-\frac{K}{E}} = \frac{1}{p(K)} - 1$$
$$\iff p(K) = \frac{1}{1 + e^{-\frac{K}{E}}} \qquad (16)$$

It shows that the performance function is actually identical to the type of sigmoid function that classical feedforward network architectures very often use as threshold function. The traditional explanation for the use of this particular threshold function has always lain in its features of non-linearity and simplicity. Non-linearity increases the expressiveness of a neural network, allowing decision boundaries in feature space that a simple linear network would not be able to model. A neural network with only linear output functions would simply collapse into a single linear function, which cannot model complex decision boundaries. The other advantage of the sigmoid function in (16) is the simplicity of its derivation, which facilitates the backpropagation of errors during the training of neural networks. While these are surely important points, it now seems that the deeper meaning of the sigmoid function has more of an information-theoretical nature, as motivated above.

Figure 3 shows the sigmoid function in (16) for four different parameters $E$, namely $E = 1$, $E = \frac{1}{2}$, $E = \frac{1}{3}$, and $E = 2$. As its name already suggests, the sigmoid function has an S-shape. It converges on 0 towards negative infinity and on 1 towards infinity. The parameter $E$ controls the steepness of the sigmoid function. For smaller values of $E$, the sigmoid function becomes steeper and approaches faster to either 0 or 1 on both ends. Independent of $E$, the sigmoid function is always 0.5 for $K = 0$.

## 6 The Golden Ratio

I now assume that the performance of a given confidence value $K$ always matches exactly the expectation, i.e. in other words $E = p(K)$. Note that this corresponds to the form of the summands of the general entropy in Section 2. The net force equation in (15) will

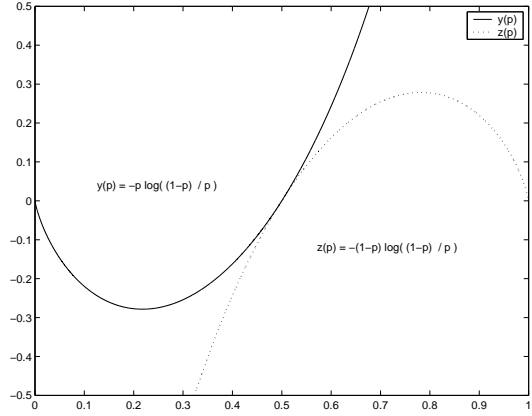

Figure 4: Net Force.

then read as follows:

$$K = -p(K) * \ln\left(\frac{1 - p(K)}{p(K)}\right) \qquad (17)$$

Figure 4 depicts the net force in (17) graphically for performance values $p(K)$ ranging from 0 to 1. As we can see in Figure 4, the net force becomes zero for $p(K) = 0$ and $p(K) = 0.5$. For performances higher than 0.5 and approaching 1, the net force diverges to infinity. Figure 4 also shows a mirrored variant of the net force, namely

$$K = -(1 - p(K)) * \ln\left(\frac{1 - p(K)}{p(K)}\right) \qquad (18)$$

This equation is a direct result of (17) after changing the sign and replacing the performance $p(K)$ with its complement $1 - p(K)$. The net force and its mirrored variant both meet at $p(K) = 0.5$. We can actually consider $p(K) = 0.5$ as a transition point where the net force transforms into its mirrored variant. After the transition, we are still looking at the same problem. However, our point of view has changed and is now reflected by the mirrored net force. This will become important later in Section 7, where we relate these forces to Yin and Yang.

For the time being, let us concentrate on the net force in (17). The net force and the counter-confidence (Force B) in (13), with $E = p(K)$, become equal when the performance $p(K)$ satisfies the following relationship:

$$p(K) = \frac{1 - p(K)}{p(K)} \qquad (19)$$
$$\iff p(K)^2 + p(K) - 1 = 0$$
$$\iff p(K) = -\frac{1}{2} \pm \sqrt{\frac{1}{4} + 1}$$
$$\iff p(K) = \frac{\sqrt{5} - 1}{2} \vee \frac{-1 - \sqrt{5}}{2}$$
$$\iff p(K) \approx 0.618 \vee -1.618 \qquad (20)$$

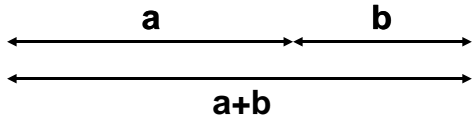

**a+b** is to **a** as **a** is to **b**

Figure 5: Golden Ratio.

This transformation shows that counter-confidence and net force are the same for a performance of about 0.618, when just considering the positive performance value. Interestingly, this transformation also shows that the two possible values satisfying (19), namely $\approx -1.618$ and $\approx 0.618$, are precisely the negative values of the so-called golden ratio. Force B thus equals the compound effect of Force A and Force B for performances defined by the golden ratio.

With a detailed introduction into the golden ratio being out of scope, I provide only some background information about the golden ratio, or golden mean as it is also called [10, 2]. The golden ratio is an irrational number, or rather two numbers, describing the proportion of two quantities. Expressed in words, two quantities are in the golden ratio to each other, if the whole is to the larger part as the larger part is to the smaller part. The whole in this case is simply the sum of both parts. Figure 5 shows an example of a line divided into two segments that are in the golden ratio to each other. Historically, the golden ratio was already studied by ancient mathematicians. It plays an important role in different fields like geometry, biology, physics, and others. Many artists and designers deliberately or unconsciously make use of it because it seems that artwork based on the golden ratio has an esthetic appeal, and features some kind of natural symmetry. Despite the fact that the golden mean is of paramount importance to so many fields, I think it is fair to say that we still do not have a full, or rather correct, understanding of its true meaning in science. Mathematically, the golden mean can be derived from the following equation, which describes the colloquial description given above in mathematical terms.

$$\frac{a+b}{a} = \frac{a}{b} \qquad (21)$$

Accordingly, the golden mean, which is typically denoted by the Greek letter $\varphi$, is then given by the ratio of $a$ and $b$, i.e. $\varphi = \frac{a}{b}$. Using the relationship in (21), the golden ratio $\varphi$ can be resolved into two possible values:

$$\varphi = \frac{1+\sqrt{5}}{2} \vee \frac{1-\sqrt{5}}{2} \qquad (22)$$

$$\implies \varphi \approx 1.618 \vee -0.618 \qquad (23)$$

Usually, the positive value ($\approx 1.618$) is identified with $\varphi$. Note that these values are the same as in (20), except that their signs are reversed. The reader interested in a thorough analysis of the golden mean can find more information and many practical examples in the references [10, 2].

# 7 Yin and Yang

I will now relate the above theoretical results with one of the oldest philosophical world views, namely the principle of Yin and Yang. In particular, I dare to advance the hypothesis that both Force A and Force B, which I defined respectively in (12) and (13) using fixed point equations, correspond to the two opposing forces Yin and Yang when we assume that expectation always equals performance, i.e. $E = p(K)$. If this can indeed be confirmed by further observations, this ancient philosophical concept could play an important role in computer science. In fact, I will provide further evidence of this claim and also show how we can use the concept of Yin and Yang for machine learning. Let me begin with a short summary of the Yin/Yang concept in Chinese philosophy.

## 7.1 Philosophy

The concept of Yin and Yang is deeply rooted in Chinese philosophy [23]. Its origin dates back at least 2500 years, probably much earlier, playing a crucial role in the oldest Chinese philosophical texts. Chinese philosophy has attached great importance to Yin/Yang ever since. Today, the idea of Yin/Yang pervades fields as different as religion, sports, medicine, politics, and many more. The fact that the Korean national flag sports a Yin/Yang symbol illustrates the emphasis laid on this concept in Asian countries.

Yin and Yang stand for two principles that are opposites of each other, and which are constantly trying to gain the upper hand over each other. However, neither one will ever succeed in doing so, though one principle may temporarily dominate the other one. Both principles cannot exist without each other. It is rather the constant struggle between both principles that defines our world and produces the rhythm of life. According to Chinese philosophy, Yin and Yang are the foundation of our entire universe. They flow through, and thus affect, every being. Typical examples of Yin/Yang opposites are, for example, night/day, cold/hot, rest/activity, etc.

Chinese philosophy does not confine itself to a mere description of Yin and Yang. It also provides guidelines on how to live in accordance with Yin and Yang. The central statement is that Yin and Yang need to be in harmony. Any imbalance of an economical, biological, physical, or chemical system can be directly attributed to a distorted equilibrium between Yin and Yang. For instance, an illness accompanied by fever is the result of Yang being too strong and dominating Yin. On the other hand, dominance of Yin could result, for instance, in a body shivering with cold. The optimal state every being, or system, should strive for is therefore the state of equilibrium between Yin and Yang. It is this state of equilibrium between Yin and Yang that Chinese philosophy considers the most powerful and stable state a system can assume.

Yin and Yang can be further subdivided into Yin and Yang. For instance, "cold" can be further divided into "cool" or "chilly," and "hot" into "warm" or "boiling." Yin and Yang already carry the seed of their opposites: A dominating Yin becomes susceptible to Yang and will eventually turn into its opposite. On the other hand, a dominating Yang gives rise to Yin and will thus turn into Yin over time. This defines the perennial alternating cycle of Yin or Yang dominance. Only the equilibrium between Yin and Yang is able to overcome this cycle.

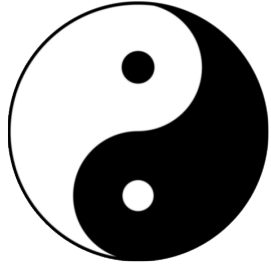

Figure 6: Yin and Yang.

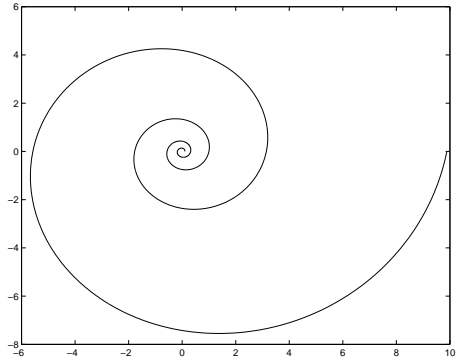

Figure 7: Logarithmic spiral.

## 7.2 Logarithmic Spirals

Figure 6 depicts the well-known black and white symbol of Yin and Yang. The dots of different color in the area delimited by each force symbolize the fact that each force bears the seed of its counterpart within itself. According to the principle of Yin and Yang outlined above, neither Yin nor Yang can be observed directly. Both Yin and Yang are intertwined forces always occurring in pairs, rather than being isolated forces independent from each other. In Chinese philosophy, Yin and Yang assume the form of spirals. I will now show that the net force in (17) is a spiral too. In order to do so, I will first introduce the general definition of the logarithmic spiral before I then illustrate the similarity to the famous Yin/Yang symbol.

A logarithmic spiral is a special type of spiral curve, which plays an important role in nature. It occurs in all different kinds of objects and processes, such as mollusk shells, hurricanes, galaxies, and many more [1]. In polar coordinates $(r, \theta)$, the general definition of a logarithmic spiral is

$$r = ae^{b\theta} \qquad (24)$$

Parameter $a$ is a scale factor determining the size of the spiral, while parameter $b$ controls the direction and tightness of the wrapping. For a logarithmic spiral, the distances between the turnings increase. This distinguishes the logarithmic spiral from the Archimedian spiral, which features constant distances between turnings. Figure 7 depicts a typical example of a logarithmic spiral. Resolving (24) for $\theta$ leads to the following general

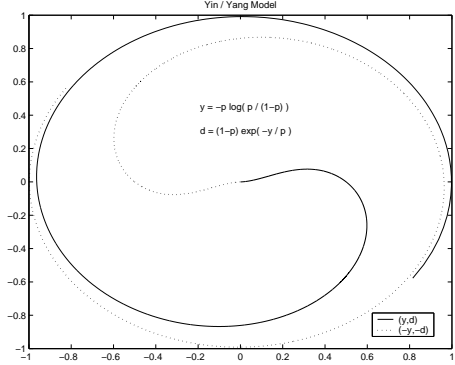

Figure 8: Yin-Yang Spirals.

form of logarithmic spirals:

$$\theta = \frac{1}{b} \ln\left(\frac{r}{a}\right) \qquad (25)$$

In order to show that the net force in (17) defines a logarithmic spiral, and for the sake of easier illustration, I investigate the negative version of the net force in (17) and look at the polar coordinates $(r, \theta)$ it defines, namely:

$$\theta = -p(K) * \ln\left(\frac{p(K)}{1 - p(K)}\right) \quad \text{and} \quad r = (1 - p(K)) * e^{\frac{-\theta}{p(K)}} \qquad (26)$$

A comparison of (26) with the general form of logarithmic spirals in (25) shows that the net force does indeed describe a spiral. Both (25) and (26) match when we set the parameters $a$ and $b$ to the following values:

$$a = 1 - p(K) \quad \text{and} \quad b = -\frac{1}{p(K)} \qquad (27)$$

In particular, we can check that $a$ and $b$ are identical when $p(K)$ equals the golden ratio. If we let $p(K)$ run from 0 to 1, and mirror the resulting spiral along both axes similar to Figure 4, we receive two spirals. Figure 8 shows both spirals plotted in a Cartesian coordinate system. Both spirals are, of course, symmetrical and their turnings approach the unit circle. A comparison of the Yin/Yang symbol of Figure 6 with the spirals in Figure 8 shows the strong similarities between both figures. A simple mirror operation transforms the spirals in Figure 8 into the Yin/Yang symbol.

The addition of a time dimension to Figure 8 generates a three-dimensional object. It resembles a funnel or trumpet that has a wide circular opening on the upper end and narrows towards the origin. Figure 9 depicts this "informational universe," which follows directly from the two-dimensional graphic in Figure 8 when I use the performance values as time coordinates for the third axis. Note that the use of performance as time is reasonable because the exponential distribution is typically used to model dynamic time processes and the expectation value is thus typically associated with time. This will also be an important point in the next section.

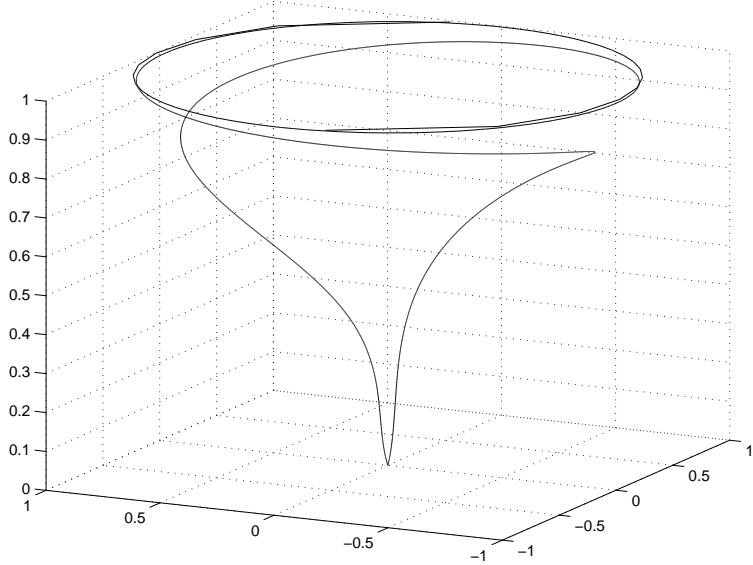

Figure 9: Informational Universe.

## 8 Relativity

This section discusses the net force in a wider context and from a physical point of view. I begin by revisiting the net force as introduced in (15):

$$K = -E * \ln\left(\frac{1-p(K)}{p(K)}\right) \qquad (28)$$

The net force describes the net effect of the two forces defined in (12) and (13), respectively. As I showed above, each force entails its own interpretation of the performance function $p(K)$. However, the net effect of both forces in (28), which computes simply as the difference between both forces, provides no information about the interpretation of $p(K)$. Both interpretations, i.e. the exponential distribution or its complement, are valid performances. In fact, the interpretation we use depends on our viewpoint and just changes the sign of the net force in (28). The previous result in (16) shows, that the sigmoid function provides the correct performance values once we have chosen our point of view. Accordingly, the performance will lie between 0 and 0.5 for a negative net force and between 0.5 and 1 for a positive net force. The fact that there is no objectively correct viewpoint strongly resembles the principle of relativity, which plays a major role in physics.

Motivated by the general entropy introduced at the beginning of this paper, I will now derive another interesting result relating to relativity. As I have introduced in Section 2, the general entropy is based on summands having the following form:

$$K_i = -p(K_i)\ln\left(p(K_i)\right) \qquad (29)$$

We can easily see that each summand matches the definition of Force B introduced in (13) when the expectation equals the performance. For this reason, I consider Force B, or rather general entropy, to be the more fundamental force of both Force A and Force B. Actually, I understand that the difference between Force A and Force B, i.e. the net force, describes merely our perception, while the general entropy defines the true uncertainty. The sigmoid function will thus provide the real performance values, allowing us to compute the actual general entropy. Spinning this thought further, I understand that we perceive reality in points defined by the golden ratio. Our perception will be different from reality except for performance values equal to the golden ratio. Let me present an interesting physical application of this idea: In physics, a typical performance function could be the velocity $v$ of an object in relation to light speed $c$. This value should always lie within the range from 0 to 1 because the current state-of-the-art assumes that no object can move faster than the speed of light. If we insert this relative speed into (19), which describes the relationship defining the golden ratio, we obtain the following result:

$$p(K) = \frac{1 - p(K)}{p(K)}$$
$$\implies \sqrt{1 - p(K)^2} = \sqrt{p(K)}$$
$$\implies \sqrt{1 - \left(\frac{v}{c}\right)^2} = \sqrt{p(K)} \tag{30}$$

The expression on the left-hand side is the well-known Lorentz factor, or rather the inverse Lorentz factor, which plays a crucial part in Einstein's special relativity. The Lorentz factor describes how mass, length, and time change for an object or system whose velocity approaches light speed. For a moving object, an observer will measure a shorter length, more mass, and a shorter time lapse between two events. These effects become more pronounced as the moving object approaches the speed of light. Depending on the relative speed to light, the Lorentz factor describes basically the ratio between the quantity measured for the observer and the quantity measured for the moving system. For instance, if $t$ is the time measured locally by the observer, then the corresponding time $t'$ measured for the moving system computes as follows:

$$t' = \sqrt{1 - \frac{v^2}{c^2}} * t \tag{31}$$

We can see that $t'$ converges to zero for increasing speed, i.e. we can measure no time lapse for a system moving with light speed. Similar relationships hold for length and mass. However, time dilation is especially interesting because the exponential distribution is very often used to model the time between statistical events that happen at a constant average rate, such as radioactive decay or the time until the next system failure, as already mentioned in the previous section. The expectation value of the exponential distribution is then indeed time, namely the expected time until the next event. In this context, an expectation value in the form of the Lorentz factor makes perfect sense. Actually, time dilation can then be followed from the relationship in (30). However, according to the perceptual model introduced above, I understand that time dilation is merely our perception and does not reflect reality. The true performance follows when we use our observed performance as input to the sigmoid function, which then provides the actual performance. For instance, an expectation value corresponding to a Lorentz factor with a relative speed (performance) of 0.5 leads to an observed performance of $\frac{1}{\sqrt{2}}$ according to (30). Insertion of this observed performance into the sigmoid function leads to the following result:

$$\frac{1}{1 + \frac{1}{\sqrt{2}}} \approx 0.586 \tag{32}$$

Note that this result is slightly larger than 0.5.

This concludes my theoretical foray into the field of physics. We know from practical experiments that the observation of a physical experiment can actually change its outcome. The classic example for this fact is the famous double slit experiment [21]. For this reason, some physicists have already suggested that they might have to include human perception into their models in order to develop a more complete and thus more powerful theory that can describe these effects. It remains to be seen to what extent the proposed perceptive model turns out to be useful in this respect.

## 9 Informational Intelligence

In this section, I am going to apply the concept of informational confidence to a practical problem. In order to do so, I divide this section into three subsections: In the first subsection, I show how to learn informational confidence values by estimating the necessary parameters on an evaluation set. In the second subsection, I present practical recognition rates of a multiple classifier system for handwritten Japanese character recognition. In the third subsection, I propose a new framework for machine learning in the form of a network architecture that implements the ideas introduced above, in particular general entropy. I therefore use the term "informational intelligence" as the title for this section in order to convey the broader meaning of informational confidence.

### 9.1 Informational Confidence Learning

In most practical cases, classifiers do not provide informational confidence values. Their confidence values typically violate the fixed point equation in the performance theorem, indicating a distorted equilibrium between information and confidence. Classifier combination therefore calls for a second training process in addition to the classifier-specific training methods teaching each classifier the decision boundaries of each class. Accordingly, I consider learning of informational confidence values to be a 3-step process: In the first step, I train a classifier with its specific training method and training set. In the second step, I estimate the performance for each confidence value on an evaluation set. Finally, I compute new informational confidence values by inserting the performance estimates into the fixed point equation of the performance theorem. The newly computed informational confidence values are stored in a look-up table and will replace the original raw confidence values in all future classifications. The fixed point equation of the performance theorem then formulates as follows:

$$K_i^{new} = -\hat{E} * \ln\left(1 - \hat{p}(K_i^{old})\right), \tag{33}$$

where $\hat{p}(K_i^{old})$ is the performance estimate of each raw confidence value $K_i^{old}$, $\hat{E}$ is the expectation estimate, and $K_i^{new}$ is the new informational confidence value subsequently replacing $K_i^{old}$.

In the following, I show how I compute the estimates $\hat{E}$ and $\hat{p}(K_i^{old})$ on the evaluation set [3, 4, 6].

### 9.1.1 Expectation Estimate $\hat{E}$

For the practical experiments in the next subsection, the classifier's global recognition rate $R$ on the evaluation set will serve as the expectation estimate $\hat{E}$. I additionally normalize the recognition rate $R$ according to the overall information $I(C)$ provided by classifier $C$. Following the computation of information for confidence values $I(1 - p(K))$, I estimate $I(C)$ using the performance complement [3, 4]:

$$\hat{I}(C) = I(1 - R) = -\ln(1 - R), \tag{34}$$

Based on the estimate $\hat{I}(C)$, $\hat{E}$ computes as $\sqrt[\hat{I}(C)]{R}$, which maps the global recognition rate $R$ to its normalized rate for a one-bit classifier. The fixed point equation in the performance theorem now formulates as follows:

$$K_i^{new} = -\sqrt[\hat{I}(C)]{R} * \ln\left(1 - \hat{p}(K_i^{old})\right) \tag{35}$$

This leaves us with the performance estimate as the only missing parameter to compute informational confidence values.

### 9.1.2 Performance Estimate $\hat{p}(K_i^{old})$

Motivated by the performance theorem, which states that the performance function follows an exponential distribution, I propose an estimate that expresses performance as a percentage of the maximum performance possible. Accordingly, my relative performance estimate describes the different areas delimited by the confidence values under their common density function. Mathematically, the performance estimate is based on accumulated partial frequencies defined by the following formula [17, 18]:

$$\hat{p}(K_i^{old}) = \frac{\sum_{k=0}^{i} n_{correct}(K_k^{old})}{N} \tag{36}$$

In this equation, $N$ is the number of patterns contained in the evaluation set. The help function $n_{correct}(K_k^{old})$ returns the number of patterns correctly classified with confidence $K_k^{old}$. The use of monotonously increasing frequencies guarantees that the estimated informational confidence values will not affect the order of the original raw confidence values:

$$K_i^{old} \leq K_j^{old} \implies K_i^{new} \leq K_j^{new} \tag{37}$$

For this reason, the performance estimate in (36) ensures that informational confidence values have no affect on the recognition rate of a single classifier, except for ties introduced by mapping two different confidence values to the same informational confidence value.

Ties can happen when two neighboring confidence values show the same performance and become indistinguishable due to insufficient evaluation data. In most applications, this should be no problem, though. Typically, the effect of informational confidence values shows only when we combine several classifiers into a multiple classifier system, with all classifiers learning their individual informational confidence values, unless we compute class-specific informational confidence values.

Estimates based on accumulated partial frequencies act like a filter in that they do not consider single confidence values but a whole range of values. They average the estimation error over all confidence values in a confidence interval. This diminishes the negative effect of inaccurate measurements of the estimate $\hat{p}(K_i^{old})$ in application domains with insufficient or erroneous evaluation data. Furthermore, estimation of informational confidence values can be considered a warping process aligning the progression of confidence values with the progression of performance. For experiments with other possible performance estimates, readers are referred to the references [3, 4, 6].

After normalization of the performance estimate $\hat{p}(K_i^{old})$ to a one-bit classifier, as I already did for the expectation estimate, the final version of the fixed point equation in the performance theorem reads as follows:

$$K_i^{new} = - \sqrt[\hat{I}(C)]{R} * \ln\left(1 - \sqrt[\hat{I}(C)]{\hat{p}(K_i^{old})}\right) \tag{38}$$

Note that the newly computed informational confidence values $K_i^{new}$ are an attractor of this fixed point equation. In other words, the fixed point will be reached after exactly one iteration of the training procedure, or rather estimation process. All additional iterations will produce exactly the same confidence values; i.e., $K_i^{new} = K_i^{old}$.

## 9.2 Practical Experiments

In this mainly theoretical paper, I confine myself to practical experiments for a multiple classifier system developed to recognize handwritten Japanese characters. Readers will find more information in the references, including other experiments with informational confidence values for document processing applications [3, 4, 6]. Handwriting recognition is a very promising application field for classifier combination. Multiple classifier systems have therefore a long tradition in handwriting recognition [22, 20]. In particular, the duality of handwriting recognition, with its two branches off-line recognition and on-line recognition, makes it suitable for multiple classifier systems. While off-line classifiers process static images of handwritten words, on-line classifiers operate on the dynamic data and expect point sequences over time as input signals. Compared to the time-independent off-line representations used by off-line classifiers, on-line classifiers suffer from stroke-order and stroke-number variations inherent in human handwriting and thus in on-line data. On the other hand, on-line classifiers are able to exploit the dynamic information and can very often discriminate between classes with higher accuracy. Off-line and on-line classifiers thus complement each other, and their combination can overcome the problem of stroke-order and stroke-number variations. This is especially important in Japanese and Chinese character recognition because the average number of strokes per character, and thus the number of variations, is much higher than in the Latin alphabet [5, 9].

| Japanese | offline | online | AND | OR |
|---|---|---|---|---|
| 1-best | **89.94** | 81.04 | 75.41 | 95.56 |
| 2-best | 94.54 | 85.64 | 82.62 | 97.55 |
| 3-best | 95.75 | 87.30 | 84.99 | 98.06 |

Table 1: Single n-best rates for handwritten Japanese character recognition.

For my experiments, I use a multiple classifier system comprising two classifiers for on-line handwritten Japanese characters. Both classifiers are nearest neighbor classifiers. One of these two classifiers, however, transforms the captured on-line data into an off-line pictorial representation before applying the actual classification engine. This transformation happens in a pre-processing step and connects neighboring on-line points using a sophisticated painting method [19, 7]. We can therefore consider this classifier to be an off-line classifier. As mentioned above, learning of informational confidence values is a three-step process: First, each classifier is trained with its standard training method on a given training set. Then, I compute the performance of each confidence value for each classifier on an evaluation set, using the performance estimate in (36). In the last step, I estimate the informational confidence values based on the estimate given in (38). These estimates will then replace the original confidence values in all future classifications of each classifier. In my experiments, each classifier was initially trained on a training set containing more than one million handwritten Japanese characters. The test and evaluation set contains 54,775 handwritten characters. From this set, I take about two third of the samples to estimate the performances of confidence values and one third to compute the final recognition performance of the estimated informational confidence values. For more information about the classifiers and data sets used, I refer readers to the references [7, 11, 12].

Table 1 lists the individual recognition rates for the off-line and on-line classifier. It shows the probabilities that the correct class label is among the n-best alternatives having the highest confidence, with $n = 1, 2, 3$. The off-line recognition rates are much higher than the corresponding on-line rates. Clearly, stroke-order and stroke-number variations are largely responsible for this performance difference. They complicate considerably the classification task for the on-line classifier. The last two columns of Table 1 show the percentage of test patterns for which the correct class label occurs either twice (AND) or at least once (OR) in the n-best lists of both classifiers. The relatively large gap between the off-line recognition rates and the numbers in the OR-column suggests that on-line information is indeed complementary and useful for classifier combination.

Table 2 shows the recognition rates for combined off-line/on-line recognition, using sum-rule, max-rule, and product-rule as combination schemes. Sum-rule adds the confidence values provided by each classifier for the same class, while product-rule multiplies the confidence values. Max-rule simply takes the maximum confidence without any further operation. The class with the maximum overall confidence will then be chosen as the most likely class for the given test pattern. Note that sum-rule is the mathematically appropriate combination scheme for integration of information from different sources [15].

| Japanese (**89.94**) | Raw Confidence | Inf. Confidence |
|:---:|:---:|:---:|
| Sum-rule | 93.25 | **93.78** |
| Max-rule | 91.30 | 91.14 |
| Product-rule | 92.98 | 65.16 |

Table 2: Combined recognition rates for handwritten Japanese character recognition.

In addition, sum-rule is robust against noise, as was shown in [8]. The upper left cell of Table 2 lists again the best single recognition rate from Table 1, achieved by the off-line recognizer. The second column contains the combined recognition rates for the raw confidence values as provided directly by the classifiers, while the third column lists the recognition rates for informational confidence values computed according to (38). Compared to the individual rates, the combined recognition rates in Table 2 are clear improvements. The sum-rule on raw confidence values already accounts for an improvement of almost 3.5%. The best combined recognition rate achieved with normalized informational confidence is 93.78%. It outperforms the off-line classifier, which is the best individual classifier, by almost 4.0%. Sum-rule performs better than max-rule and product-rule, a fact in accordance with the results in [8].

### 9.3  Neural Network Architecture

At the end of this paper, I am going to show how the results introduced above can be combined to form a network architecture for complex decision problems. The architecture I propose is similar to the well-known feedforward type of artificial neural networks in that a neuron first integrates its inputs and then applies a sigmoid function to compute the final output, which it propagates to the synapses of other neurons. The main motivation for the sigmoid function, however, derives from an informational-theoretical background, as discussed in Section 5. Figure 10 shows the basic unit of the proposed "information network:" a neuron and its synapses. The basic idea is that each synapse computes one summand of the general entropy defined in (3) of Section 2. The main body of the neuron first integrates all these summands, computing the general entropy according to (1) and (3). The sigmoid function then computes the actual performance based on the general entropy. Finally, the neuron forwards the newly computed performance to other neurons, which in turn repeat the same process. In this way, complex decisions become aggregates of simpler decisions.

Similar to the training process in feedforward networks, the backpropagation of feedback trains the network in Figure 10. Instead of the gradient descent in parameter space that is typically implemented in feedforward networks, backpropagation for the network in Figure 10 means basically propagating the performance back so that each neuron can adjust its output. The performance can be directly inserted as part of the sigmoid function in (16). For instance, insertion of the performance values defined in (36) leads to the following expression for the output values, after additionally normalizing each

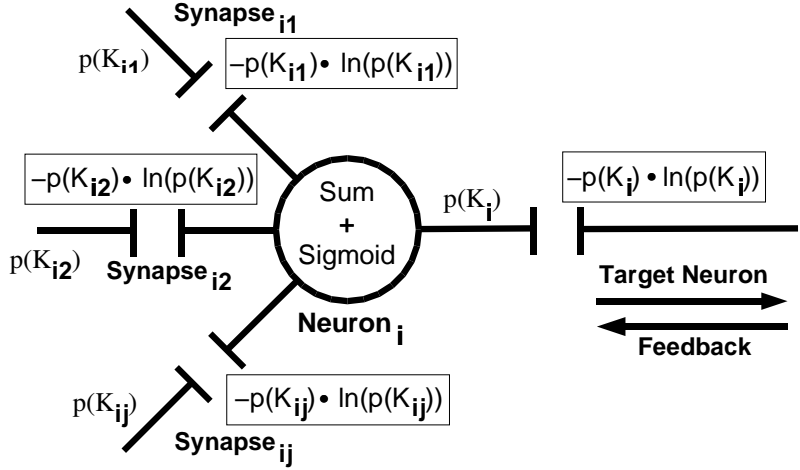

Figure 10: Information Network.

performance value to one bit:
$$\frac{1}{1 + \sqrt[\hat{I}(C)]{\hat{p}(K_i^{old})}} \tag{39}$$

In my experiments, a simple summation of the information provided by each output value, or rather classifier, for each class provides a recognition rate of 93.92 for the handwritten character recognition problem. This is better than the best recognition rate in Table 2.

I hope to be able to support the proposed network architecture with additional experiments in other application domains, and by implementing a full-fledged network and not just a single layer.

## 10  Summary

I introduced a new form of entropy that can be considered an extension of the classic entropy introduced by Shannon. Each summand of this entropy is a fixed point equation in which the so-called performance function takes over the part of the probability. However, the performance function plays several roles in my approach: It describes the distribution of an exponentially distributed random variable, and is also an expectation value in the statistical sense. Furthermore, with the exponential distribution typically used to describe statistical time processes, there is also a point in favor of it being time. The performance theorem in the first part of the paper summarizes these relationships and provides guidelines for learning informational confidence values for classifier combination. In my first practical results published in [3, 4, 6], I improved the recognition rates for several multiple classifier systems. In the present paper, I confined myself to the recognition rates for handwritten Japanese character recognition and concentrated on theoretical issues. I showed how to produce a symbol similar to the famous Yin/Yang symbol by depicting the net confidence as a spiral. The net confidence is the difference between the confidence and counter-confidence, with the latter being based on the performance complement. My understanding is that our perception is always the composite of Yin and Yang and does not reflect the reality, except when the performance function equals the golden ratio. I thus assign an information-theoretical meaning to the golden ratio. Moreover, I understand that the sigmoid function provides the actual performance value that we cannot observe directly. Under these observations and assumptions, I can explain the time dilation of Einstein's Special Relativity. However, it follows that time dilation is mere perception and does not correspond to reality. At the end of the paper, I proposed a network architecture for complex decisions, which takes advantage of the general entropy concept. I hope that the usefulness of this architecture can be confirmed by future experiments in different application fields.

## Acknowledgment

I would like to thank Ondrej Velek, Akihito Kitadai, and Masaki Nakagawa for providing data for the practical experiments.